\documentclass[conference]{IEEEtran}
\usepackage{amsmath,amsfonts}
\usepackage{algorithmic}
\usepackage{algorithm}
\usepackage{array}
\usepackage[caption=false,font=normalsize,labelfont=sf,textfont=sf]{subfig}
\usepackage{textcomp}
\usepackage{stfloats}
\usepackage{url}
\usepackage{verbatim}
\usepackage{graphicx}
\usepackage{subfig}
\usepackage{float}
\usepackage{cite}
\usepackage[numbers,sort&compress]{natbib}
\usepackage{enumerate}
\usepackage[justification=centering]{caption}
\usepackage{booktabs} 
\usepackage{multirow}
\usepackage{hyperref}
\usepackage{color}
\usepackage{caption}
\begin{document}

\title{Universal adversarial perturbation for remote sensing images}

\author{Qingyu Wang,Guorui Feng, Zhaoxia Yin, Bin Luo
\thanks{This research work is
partly supported by National Natural Science Foundation of China (62172001, U20B2068, 61860206004). (Corresponding author: Bin Luo)}
}

\author{\IEEEauthorblockN{Qingyu Wang}
\IEEEauthorblockA{\textit{School of Computer Science and Technology} \\
\textit{Anhui University}\\
Heifei, China\\
e20301182@stu.ahu.edu.cn}
\and
\IEEEauthorblockN{Guorui Feng}
\IEEEauthorblockA{\textit{School of Communication \& Electronic Engineering}\\
\textit{Shanghai University}\\
Shanghai, China\\
grfeng@shu.edu.cn}

\and

\IEEEauthorblockN{Zhaoxia Yin}
\IEEEauthorblockA{\textit{School of Communication \& Electronic Engineering} \\
\textit{East China Normal University} \\
Shanghai, China \\
zxyin@cee.ecnu.edu.cn}

\and
\IEEEauthorblockN{Bin Luo \thanks{ Bin Luo is corresponding author. }}
\IEEEauthorblockA{\textit{School of Computer Science and Technology}\\
\textit{Anhui University}\\
Heifei, China\\
luobin@ahu.edu.cn}

}




\maketitle

\begin{abstract}
Recently, with the application of deep learning in the remote sensing image (RSI) field, the classification accuracy of the RSI has been dramatically improved compared with traditional technology. However, even the state-of-the-art object recognition convolutional neural networks are fooled by the universal adversarial perturbation (UAP). The research on UAP is mostly limited to ordinary images, and RSIs have not been studied. To explore the basic characteristics of UAPs of RSIs, this paper proposes a novel method combining an encoder-decoder network with an attention mechanism to generate the UAP of RSIs. Firstly, the former is used to generate the UAP, which can learn the distribution of perturbations better, and then the latter is used to find the sensitive regions concerned by the RSI classification model. Finally, the generated regions are used to fine-tune the perturbation making the model misclassified with fewer perturbations. The experimental results show that the UAP can make the classification model misclassify, and the attack success rate of our proposed method on the RSI data set is as high as 97.09\%.
\end{abstract}

\begin{IEEEkeywords}
Remote sensing image, deep learning, universal adversarial perturbations, encoder-decoder, attention mechanism 
\end{IEEEkeywords}

\section{Introduction}
\IEEEPARstart{T}{he} development of surface observation instruments has promoted the rapid development of remote sensing image (RSI) technology and made the RSI scene classification widely used in coverage management, geographic spatial target detection, urban planning, and other research \cite{kussul2017deep,manno2015orientation}. 

Due to the small number, low resolution, and lack of diversity of RSIs, the classification accuracy of traditional feature extraction methods  \cite{oliva2001modeling,dalal2005histograms} is limited. Due to the strong learning ability of deep learning, many excellent deep algorithms have been applied to remote sensing sectors such as convolutional neural networks and achieved significant results. \cite{chaib2017deep} used the neural network to extract the image information, selected the consolidated full connection layer to construct the final RSI scene, and used discriminant correlation analysis for feature fusion. The problems of extracting space-spectral features and classification model overfitting were solved by using 3D convolutional neural networks, regularization, dropout, and virtual examples  \cite{chen2016deep}. \cite{ma2021scenenet} proposed a scene classification network architecture based on multi-objective neural evolution which can extract information of the RSI more flexible. The above classification methods baed on the neural network dramatically have improved the classification accuracy and speed of the RSI classification model. 

At present, many studies have shown that deep learning has serious security problems. Such as, the adversarial perturbation that could make the classification model misclassify clean examples was first discovered by \cite{goodfellow2014explaining}. \cite{goodfellow2014explaining} also proposed a method to generate adversarial perturbation by calculating the backpropagation values of gradients. \cite{kurakin2018adversarial} improved the attack success rate (ASR) by increasing the number of computing the gradient and reducing the attack step. \cite{moosavi2016deepfool} calculated the distance between clean examples and classification boundaries to improve the visual quality of adversarial examples. \cite{carlini2017towards} proposed an attack method based on optimization, which considered both high ASR and minor perturbation. However, the above attack methods can only calculate a single perturbation at a time, which takes a long time and only has a high ASR on the white box model with known network parameters.

Aiming at the problems existing in the above attack methods, \cite{moosavi2017universal} first proposed the universal adversarial perturbation (UAP) that image-agnostic perturbations by superposing perturbation generated by \cite{moosavi2016deepfool}. UAP is generated from multiple images and is applied to more. UAP can maintain good generalization ability that can well attack black-box models with unknown parameters. \cite{zhang2020generalizing} used spatial transformation \cite{xiao2018spatially} and  image-to-image translation adversarial networks \cite{xiao2018generating} to propose an additive and non-additive universal perturbations generation method. However, the above methods of generating UAP have the problem of poor visual quality in RSIs.


\begin{figure*}[ht]
\centerline{\includegraphics[width=14cm]{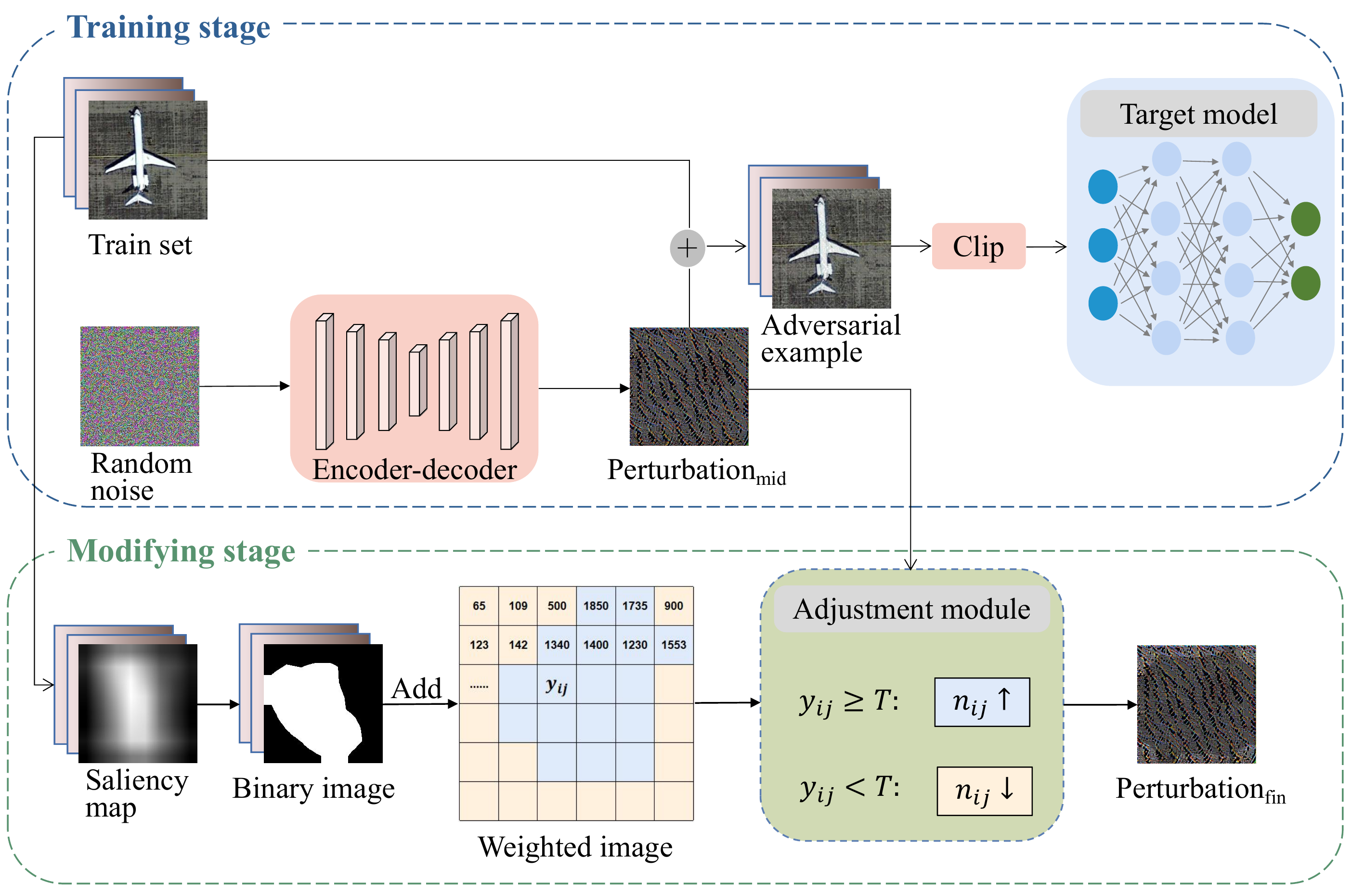}}
\caption{Overview of generalized universal adversarial perturbation. `$\uparrow$' represents that the value of $n_{ij}$ increases and `$\downarrow$' represents that the value of $n_{ij}$ decreases.
  }
\label{fig1}
\end{figure*}

Many studies found that even RSIs that differ from ordinary images in shooting angle and resolution also face the same security problems. \cite{czaja2018adversarial} first applied the attack method on ordinary images to the RSI classification model and verified that the RSI classification models can also misclassify. \cite{xu2020assessing} showed the performance of remote sensing adversarial examples under different models and data sets and found that most of the adversarial examples were wrongly divided into several specific classes, which was called attack selectivity. \cite{xu2022universal} proposed a black-box attack method using a proxy model and tested the universal of the generated adversarial examples in different models. \cite{chen2021empirical} verified that synthetic aperture radar images were also affected by adversarial attacks. \cite{wang2021universal} used the method \cite{moosavi2017universal} to verify that synthetic aperture radar images were also affected by UAP. The above research shows that adversarial attacks can cause misclassification of the RSI classification model. However, the UAP in remote sensing images has not been studied.

To verify the effect of the UAP on the RSI classification model and whether remote sensing universal adversarial examples has the attack selectivity, a method of generating remote sensing UAP using an encoder-decoder network \cite{badrinarayanan2017segnet} is proposed. To better find the characteristics of the regions concerned by the RSI classification models, an saliency map \cite{selvaraju2017grad} is used. The experimental results of our proposed method, which can achieve an ASR of 97.09\%, show that the RSI is also affected by UAP. 

The main contributions of our paper are:

\begin{enumerate}[1.]
\item An encode-decode network, which can improve the generation efficiency, is introduced to generate the universal perturbation.

\item  A saliency map, which can find the sensitive region of the classification model, is used to modify the universal perturbation.

\item We verify that universal adversarial examples of RSI have the attack selectivity.
\end{enumerate}

\section{Our proposed method}

The perturbation generated needs to mislead the classification model accurately to generate the UAP with high ASR and minor perturbation. Based on this, we propose a method to attack accurately by training and modifying UAP. In the training stage, an encoder-decoder network and a classification model are used to train the perturbation.  An encoder-decoder network is used to ensure that the input and output are consistent. The classification model is used to update the perturbation.  To reduce the perturbation of remote sensing UAP, a saliency map is used to fine-tune the generated perturbations to improve their ASR and visual quality.

\subsection{ Training stage}
Fig. \ref{fig1} is flow charts for generating UAP. In the training stage, random noise $z \sim N(0, 1)$ is an input. The generator includes multi-layer convolution, pooling, and upsampling operations, ensuring better high-dimensional feature extraction. Perturbation is added to each clean example to get the adversarial example which is then clipped. Next, the target model is used to predict it.  For target model $C_\theta(x)$ with parameter $\theta$, the model can correctly identify the clean example \textit{x} if $C_\theta(x) = c$, where \textit{c} is the correct label of the corresponding example. When the example is added with perturbation $\delta$, the model will misclassify the example if  $C_\theta(x+ \delta) \ne c$. 
The UAP is to find a perturbation $v$ that many clean examples satisfies the formula $C_\theta(x + v) \ne c$. This paper aims to find a perturbation $v$ that misclassifies most examples. The training of universal perturbations with an encoder-decoder network can be summarized by
\begin{equation}
\underset{\theta_1 }{min}{E(x,y)}\sim D[\frac{max}{\left\|\delta  \right\|_{\infty }\leq \epsilon }L(C_{\theta }(x + \delta ), y)],
\end{equation}
where $y$ is the example label, and $L(C_{\theta }(x + \delta ), y)]$ is the loss function, i.e., cross-entropy loss. $max(L)$ is the optimization objective, that is, to find the perturbation bounded by an infinite norm that maximizes the loss function. The outer layer is the minimum formula for optimizing the encoder-decoder network with parameter $\theta_1$. When the model is fixed, we minimize the perturbation to make the model classification errors. 

The loss function used to optimize the encode-decode network in this paper is
\begin{equation}
L = -\frac{1}{N}\sum_{i}^{}\sum_{c=1}^{M}y_{ic}\log(P_{ic}), 
\end{equation}
where $N$ is the total number of examples. $M$ represents the number of categories, $y_{ic}$ = 1 if the true category of example $i$ is equal to $c$, otherwise 0, $P_{ic}$ is the predicted probability that the example $i$ belongs to category $c$.

\newcommand{\mysize}{2.8cm} 
\begin{figure*}[b]
\centering
\subfloat[airplane]{
\includegraphics[width=\mysize]{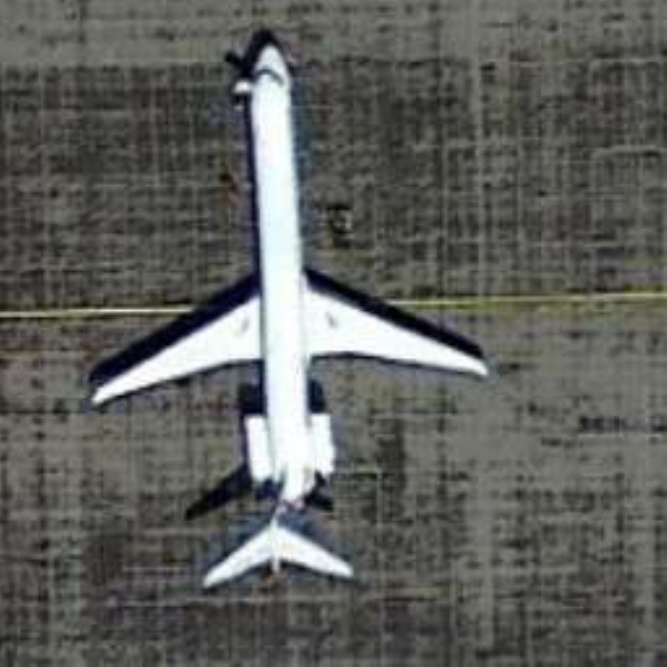}
}
\quad
\subfloat[baseball\_court]{
\includegraphics[width=\mysize]{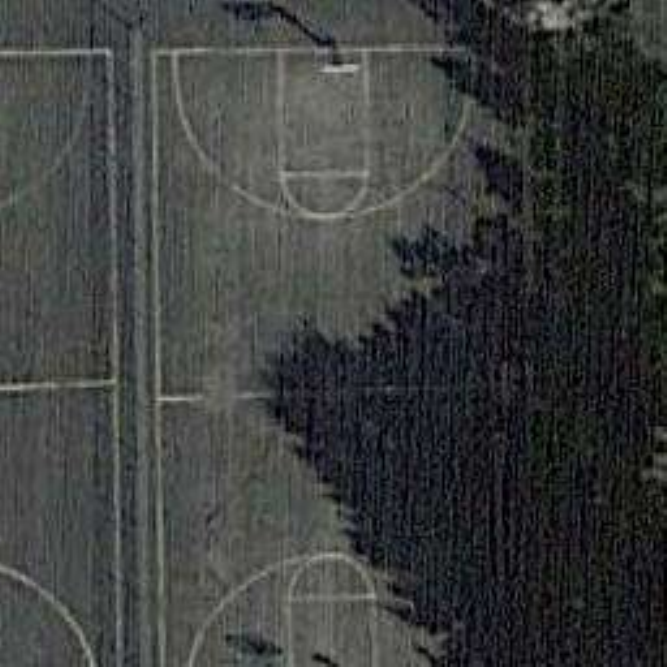}
}
\quad
\subfloat[bridge]{
\includegraphics[width=\mysize]{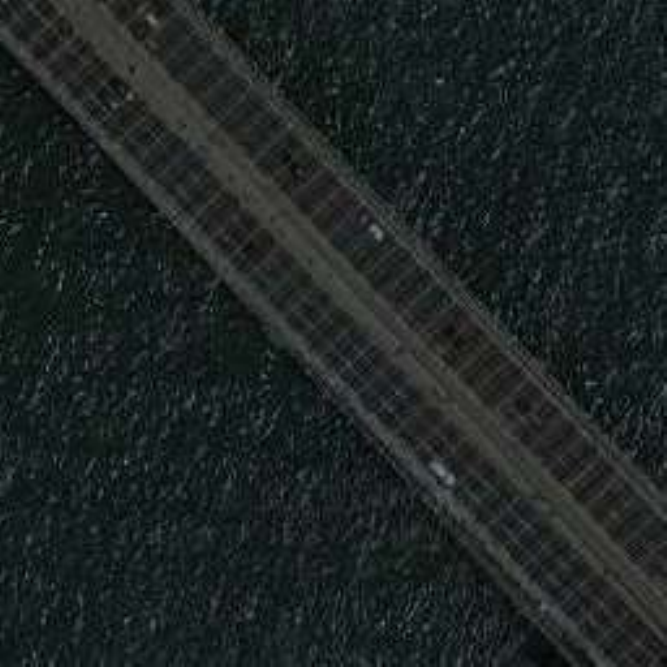}

}
\quad
\subfloat[coastal\_mansion]{
\includegraphics[width=\mysize]{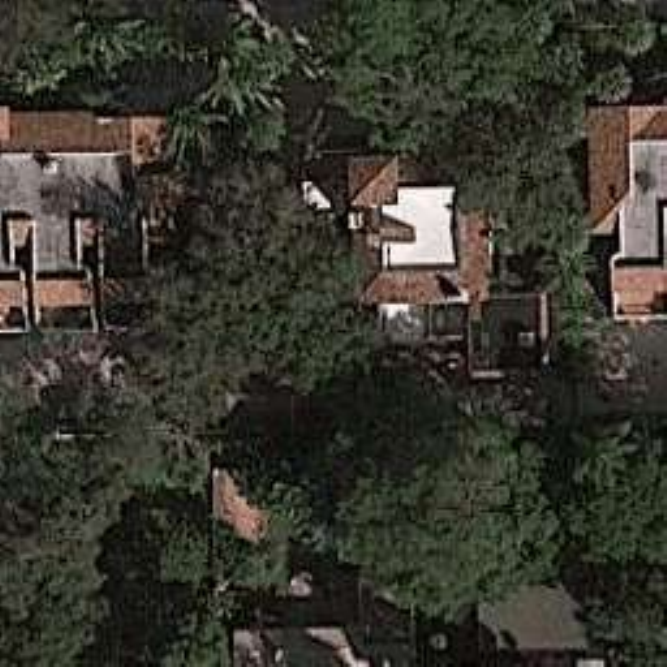}
}
\quad \\   

\subfloat[storage\_tank]{
\includegraphics[width=\mysize]{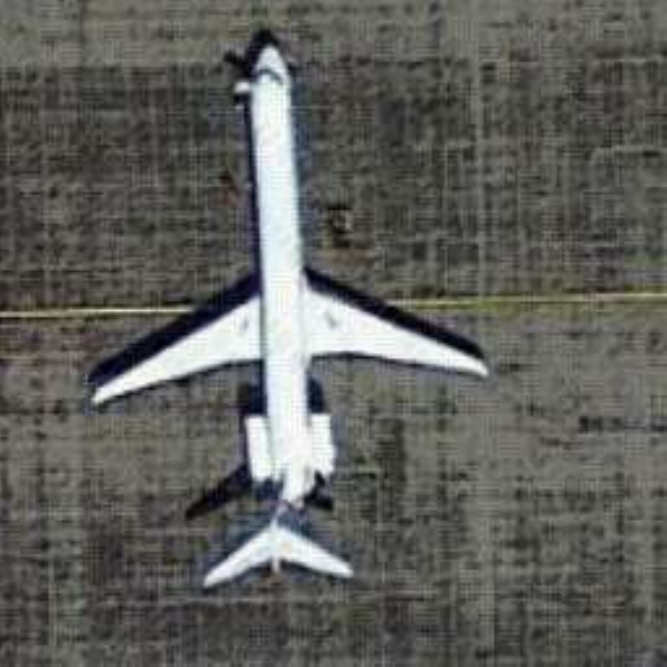}
}
\quad
\subfloat[overpass]{
\includegraphics[width=\mysize]{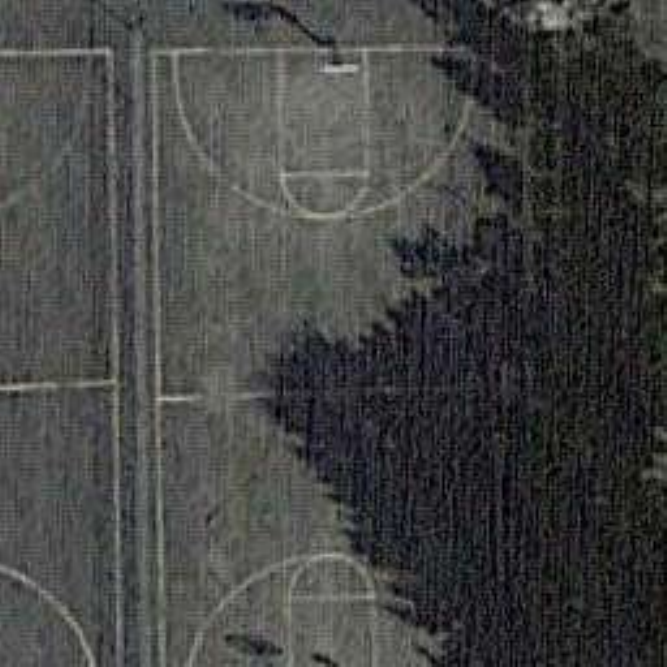}
}
\quad
\subfloat[sparse\_residential]{
\includegraphics[width=\mysize]{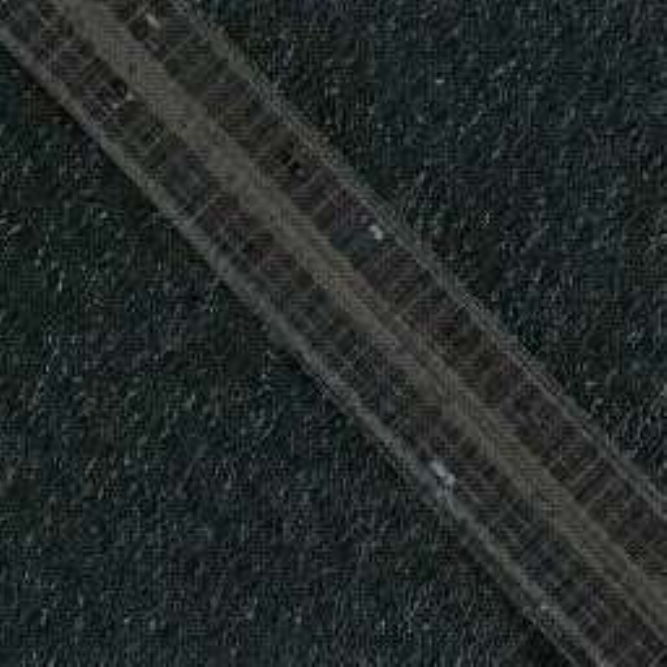}
}
\quad
\subfloat[bridge]{
\includegraphics[width=\mysize]{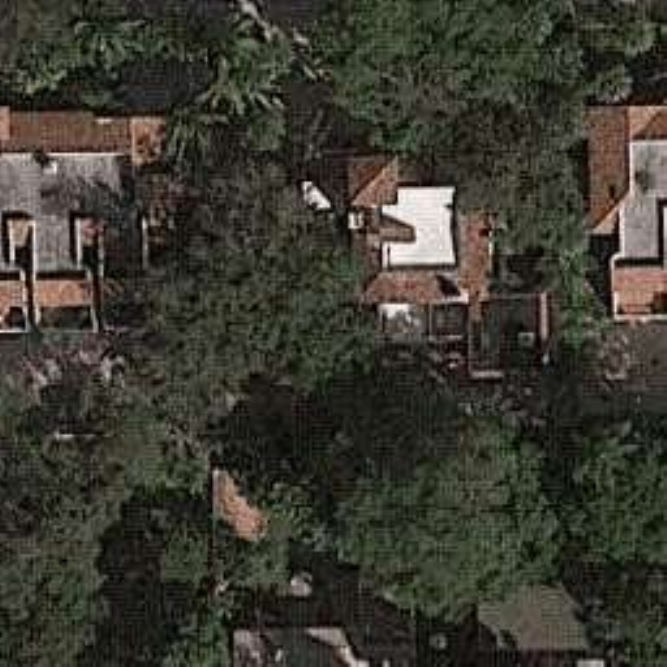}
}
\caption{Some clean and adversarial examples in VGG16. (a)$\sim$(d) are clean examples, (e)$\sim$(h) are adversarial examples.}

\label{fig2}
\end{figure*}

\subsection{Modifying stage}
In the modifying stage, firstly, we generate the saliency map, which represents the region of concern for the model of the training set. The reason why the saliency map is used to modify the perturbation is saliency map can find the concern region of the model, which is more sensitive to the perturbation than the other region. In order to better the statistical model for each sample area of concern, the saliency map is binarized. Next, a weighted image of the model concern region is obtained by adding all binary images. The weighted image is divided into two parts by setting a threshold of $T$. 

An adjustment module is designed to modify the perturbation$_{mid}$ according to the weighted image,  which is modified by
\begin{equation}
n_{ij}=\left\{
\begin{matrix}
n_{ij}*\alpha & if  \quad y_{ij} \geq T \\
n_{ij}*\beta & if \quad  y_{ij} < T
\end{matrix}
\right. ,
\end{equation}
where $y_{ij}$ represents the value  of the weighted image in position $(i,j)$ and $n_{ij}$ represents the value of perturbation$_{mid}$ in position $(i,j)$. The parameter of $\alpha$ and $\beta$ is a positive number, which is $\alpha > 1$ and $\beta < 1$. The value of the parameter is different due to the different areas of concern in different models. When the value $y_{ij}$ greater than $T$, the value $n_{ij}$ is increased; otherwise, $n_{ij}$ is decreased. Finally, the perturbation$_{fin}$ is generated, which is used to generate the universal adversarial examples.

\newcommand{\mysizew}{2.8cm} 
\begin{figure*}[ht]
\centering
\subfloat[VGG16]{
\includegraphics[width=\mysizew]{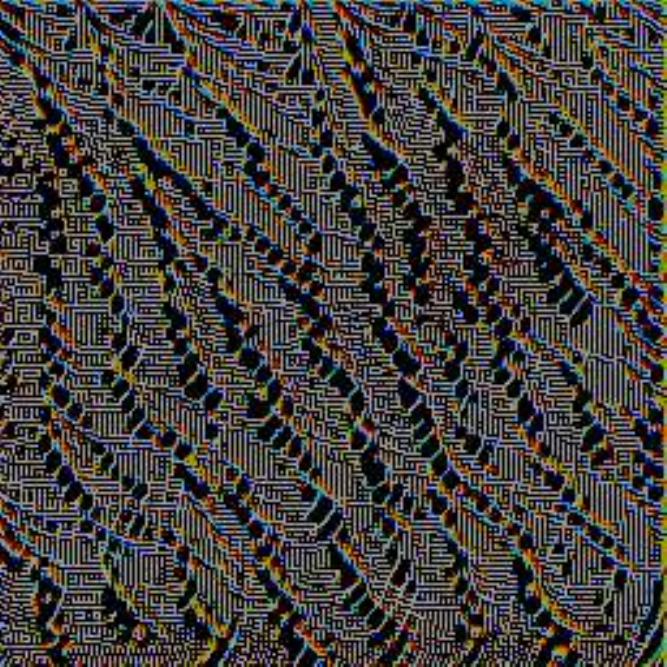}
}
\quad
\subfloat[VGG19]{
\includegraphics[width=\mysizew]{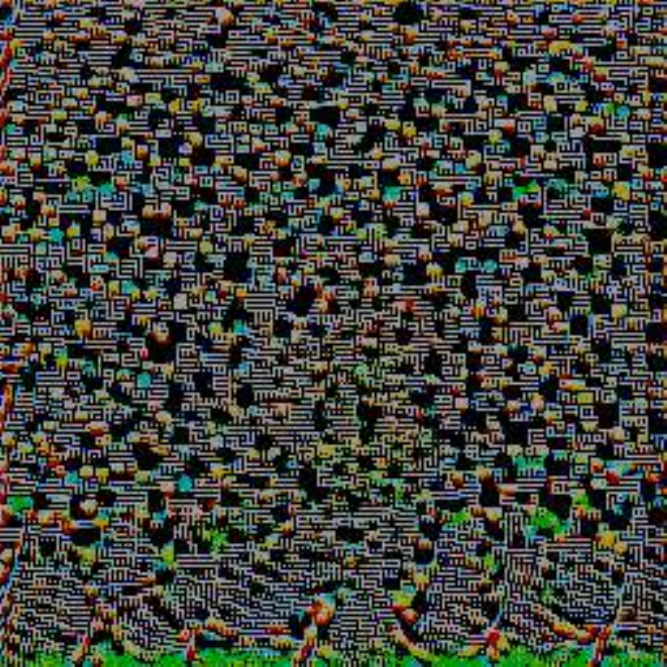}
}
\quad
\subfloat[ResNet34]{
\includegraphics[width=\mysizew]{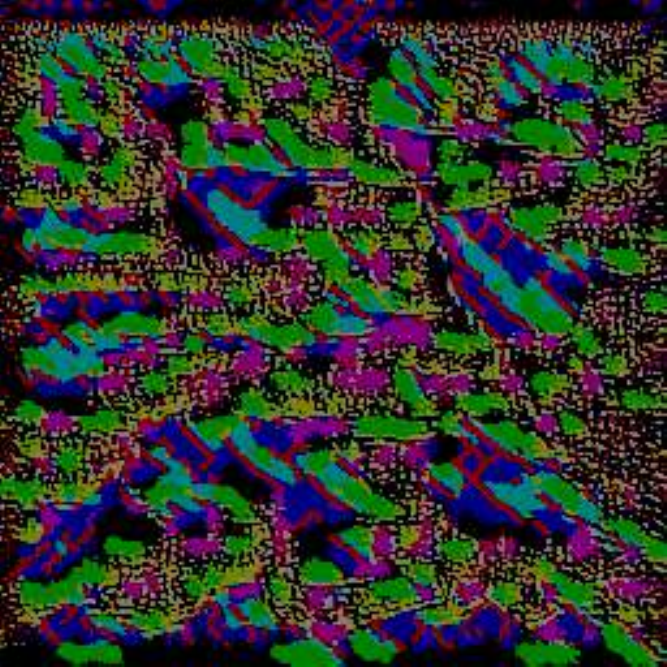}

}
\quad
\subfloat[ResNet101]{
\includegraphics[width=\mysizew]{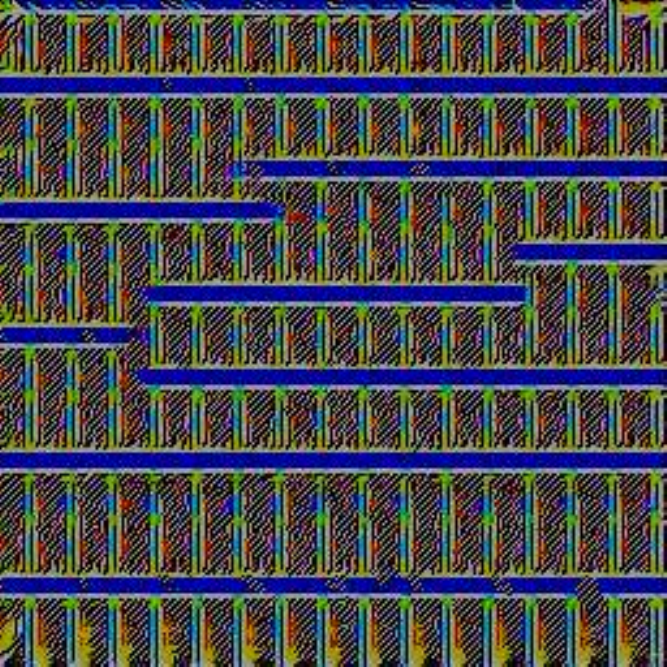}
}
\caption{Universal perturbations computed for different network architectures. The pixel values are scaled for visibility.}

\label{fig3}
\end{figure*}

\newcommand{\mysizes}{2.8cm} 
\begin{figure*}[htbp]
\centering
\subfloat{
\includegraphics[width=\mysizes]{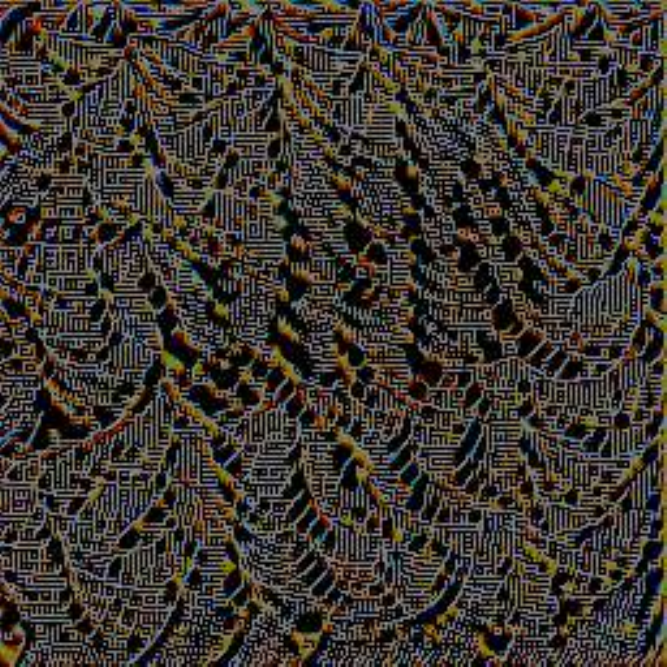}
}
\quad
\subfloat{
\includegraphics[width=\mysizes]{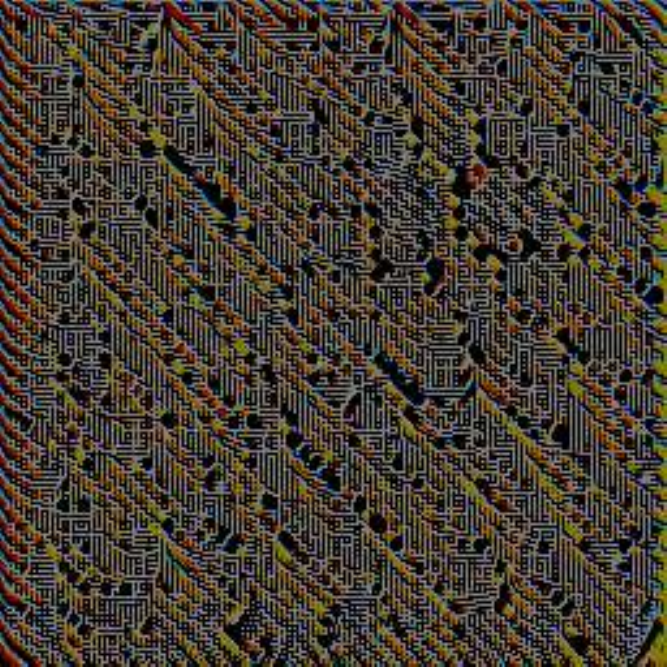}
}
\quad
\subfloat{
\includegraphics[width=\mysizes]{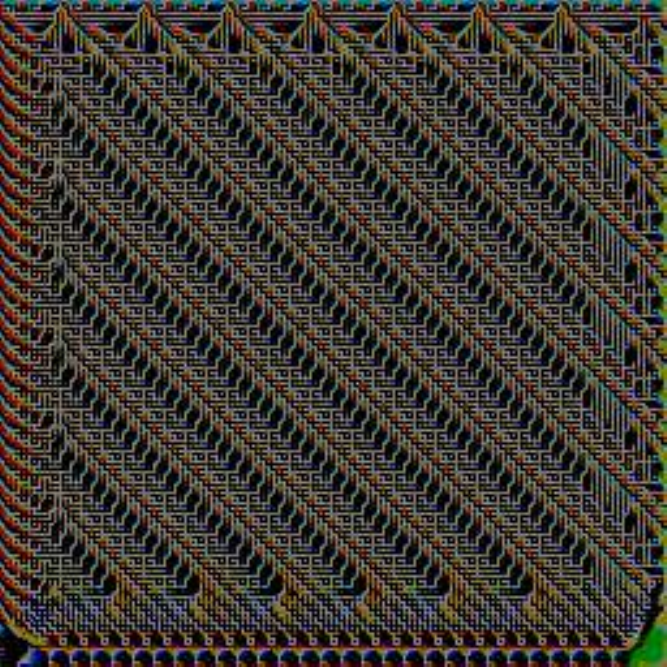}

}
\quad
\subfloat{
\includegraphics[width=\mysizes]{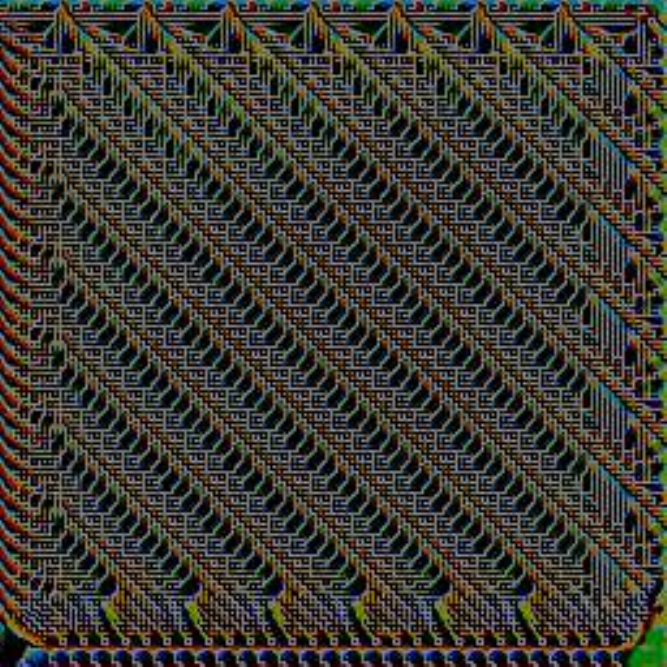}
}
\caption{Diversity of universal perturbations for the VGG16 network. The four perturbations are generated using different random shufflings of the training set.}

\label{fig4}
\end{figure*}

\section{Experments}
The experiments are carried out on the remote sensing data set PatternNet \cite{zhou2018patternnet} with 38 classes. The data set has 38 classes and each includes 800 images with a size of $256 \times 256$. We randomly select 50 images for each class as the training set and 9158 as the validation set. The latest classification models of VGG16 \cite{simonyan2014very}, VGG19 \cite{simonyan2014very}, ResNet34 \cite{he2016deep} and ResNet101 \cite{he2016deep} are trained to evaluate the attack effect of our proposed method. The classification accuracy of the four models is 94.92\%, 94.18\%, 99.72\% and 99.75\%, respectively. An infinite norm bounds the UAP in training, such as $ \left\| \delta \right\|_{\propto }\leq \varepsilon = 10$. The parameters of the learning rate and weight decay are 0.001, and the number of iterations is 50. Fig. \ref{fig2} illustrates some clean and corresponding adversarial examples. For example, the label in Fig. \ref{fig2} (a) changes from airplane to (Fig. \ref{fig2} (e)) storage\_tank when a universal perturbation is added. It can be seen that the universal adversarial perturbation is imperceptibility. The universal perturbations of different networks are visualized in Fig. \ref{fig3}. It shows that the universal perturbations are not the same in different networks. In Fig. \ref{fig4}, four different universal perturbations obtained by using random shufflings in the training set are visualized. It can be seen that such universal perturbations are different, although they are generated by the same networks and constraints.

\subsection{The performance of the ASR and perturbation magnitude (PM)}
The evaluation criteria are the ASR, which refers to the proportion of adversarial examples that the model misclassifies, and PM, which is the difference between the adversarial and clean examples. We compare ASR and PM on the validation set with the other methods. The experiment results are shown in Table \ref{tbl1}. It can find that our proposed method improves the ASR and reduces the PM. This is because an encoder-decoder network and a salience map are used to mislead the classification model accurately. In addition, we test the effectiveness of the saliency map used in our proposed method. The performance before and after perturbations modification is shown in Table \ref{tbl2}. It can be seen that the saliency map can improve the visual quality and ASR of the adversarial examples.

\begin{figure}[ht]
\centerline{\includegraphics[width=9cm]{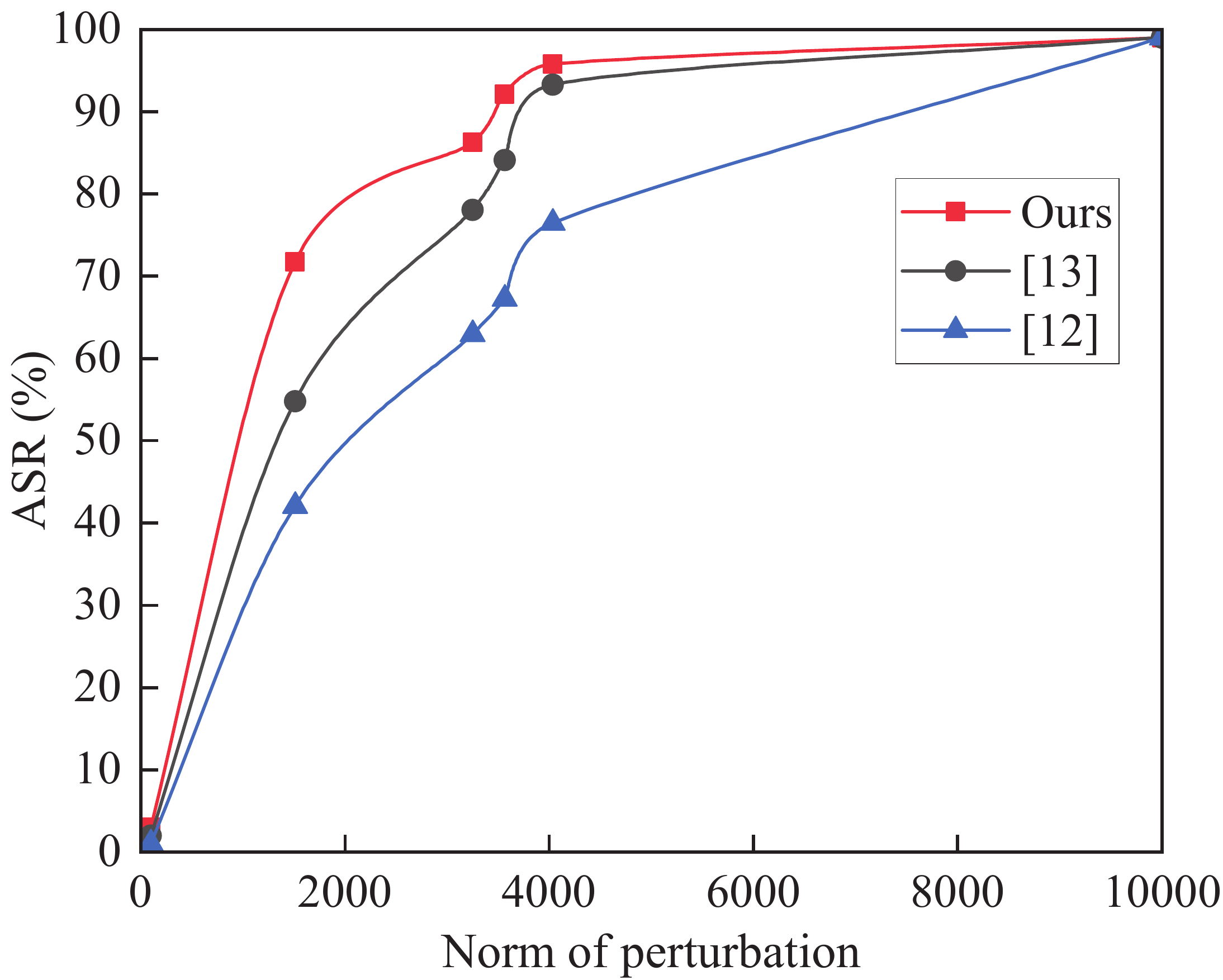}}
\caption{Comparison of ASR (\%) under different perturbations on the VGG16 network.}
\label{fig5}
\end{figure}

\renewcommand{\arraystretch}{1.2}
\begin{table}[h]
\centering
\caption{\textrm{Comparison of ASR (\%) and PM ($l_{2}$ norm) with other methods on the Patternet dataset.}}
\resizebox{\columnwidth}{!}{
\begin{tabular}{cccccc}
\toprule
 \multicolumn{2}{c}{\textrm{ }}&\textrm{VGG16}  & \textrm{VGG19} & \textrm{ResNet34} & \textrm{ResNet101}\\
\midrule
\multirow{3}{*}{ASR} & \textrm{\cite{moosavi2017universal}} & \textrm{73.27} & \textrm{66.35} & \textrm{82.91} & \textrm{89.29}\\
& \textrm{\cite{zhang2020generalizing}} & \textrm{95.40} & \textrm{93.31} & \textrm{96.22} & \textrm{96.27}\\
 & \textrm{Ours} & \textrm{\textbf{95.75}} & \textrm{\textbf{94.74}} & \textrm{\textbf{96.26}} & \textrm{\textbf{97.09}}\\
\multirow{3}{*}{PM} & \textrm{\cite{moosavi2017universal}} & \textrm{3348.64} & \textrm{3422.67} & \textrm{4015.34} & \textrm{4322.52}\\
 & \textrm{\cite{zhang2020generalizing}} & \textrm{4126.38} & \textrm{4247.44} & \textrm{4042.89} & \textrm{4209.33}\\
& \textrm{Ours} & \textrm{\textbf{3375.28}} & \textrm{\textbf{3288.22}} & \textrm{\textbf{3845.84}} & \textrm{\textbf{3878.72}}\\
\bottomrule
\end{tabular}
}
\label{tbl1}
\end{table}

\renewcommand{\arraystretch}{1.2}
\begin{table}[h]
\large
\centering
\vspace{0.2cm}
\caption{\textrm{The ablation experiment with and without saliency map  in our proposed method.}}
\resizebox{\columnwidth}{!}{
\begin{tabular}{cccccc}
\toprule
\multicolumn{2}{c}{\textrm{ }}&\textrm{VGG16} & \textrm{VGG19} & \textrm{ResNet34} & \textrm{ResNet101}\\
\midrule
\multirow{2}{*}{ASR} & \textrm{Without} & \textrm{95.72} & \textrm{94.41} & \textrm{95.82} & \textrm{97.02}\\
\textrm{} &  \textrm{With} & \textrm{\textbf{95.75}} & \textrm{\textbf{94.74}} & \textrm{\textbf{95.96}} & \textrm{\textbf{97.09}}\\
\multirow{2}{*}{PM} & \textrm{Without} & \textrm{4034.78} & \textrm{4053.13} & \textrm{3665.37} & \textrm{4159.35}\\
\textrm{} & \textrm{With} & \textrm{\textbf{3375.28}} & \textrm{\textbf{3288.22}} & \textrm{\textbf{3512.71}} & \textrm{\textbf{3878.72}}\\
\bottomrule
\end{tabular}
}
\label{tbl2}
\end{table}

\begin{table}[htbp]
\large
\centering
\caption{\textrm{Generalization ability representing the ASR (\%) in different models of our proposed method with and without saliency
map. The first row is attacked models, and the first column is the targeted models used to generate adversarial perturbations.}}
\resizebox{\columnwidth}{!}{
\begin{tabular}{cccccc}
\toprule
\multicolumn{2}{c}{\textrm{ }}&\textrm{VGG16} & \textrm{VGG19} & \textrm{ResNet34} & \textrm{ResNet101}\\
\midrule
\multirow{2}{*}{VGG16} & \textrm{Without} & \textrm{95.72 } & \textrm{92.53} & \textbf{\textrm{86.74}} & \textrm{86.74}\\
\textrm{} &                                     \textrm{With} & \textbf{\textrm{95.75}} & \textbf{\textrm{92.78}}& \textrm{85.93} & \textbf{\textrm{86.76}}\\

\multirow{2}{*}{VGG19} & \textrm{Without} & \textbf{\textrm{93.47}} & \textrm{94.41} & \textbf{\textrm{88.60}} & \textrm{83.94}\\
\textrm{} &                                     \textrm{With} & \textrm{93.46} & \textbf{\textrm{94.74}} & \textrm{88.03} & \textbf{\textrm{84.23}}\\

\multirow{2}{*}{ResNet34} & \textrm{Without} & \textbf{\textrm{76.49}} & \textbf{\textrm{73.10}} & \textrm{95.82} & \textrm{86.21}\\
\textrm{} &						   \textrm{With} & \textrm{76.10} & \textrm{72.30} & \textbf{\textrm{96.26}} & \textbf{\textrm{86.31}}\\

\multirow{2}{*}{ResNet101} & \textrm{Without} & \textbf{\textrm{84.45}} & \textbf{\textrm{82.18}} & \textrm{90.97} & \textrm{97.02}\\
\textrm{} &						 \textrm{With} & \textrm{84.22} & \textrm{81.67} & \textbf{\textrm{91.13}} & \textbf{\textrm{97.09}}\\
\bottomrule
\end{tabular}
}
\label{tbl3}
\end{table}

\begin{figure*}[b]
\centerline{\includegraphics[width=15cm]{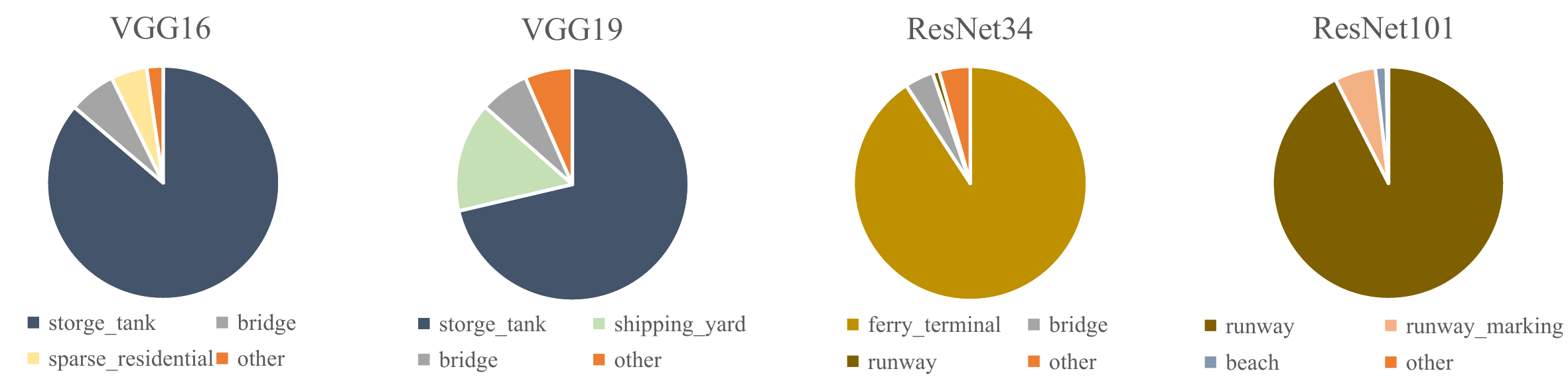}}
\caption{A demonstration of attack selectivity on the PatternNet dataset. The pie chart represents the classification result distribution of adversarial examples.}
\label{fig6}
\end{figure*}

To demonstrate the effectiveness of our method, we show the ASR under different perturbations in Fig. \ref{fig5}. These perturbations are calculated by averaged $l_{2}$ distances in the validation set. Different perturbation norms are achieved by scaling each perturbation to have the target norm. It can be seen from Fig. \ref{fig5} that the overall attack effect of our proposed method is higher compared to the other methods under the same perturbations.

\subsection{Generalization ability between models}
The generalization ability of our proposed method with and without a saliency map is tested to show the attack ability of the UAP between different models, as shown in Table \ref{tbl3}. Each row represents the ASR of the perturbation generated in the target model on the other models, and each column represents a different target model. As in Table \ref{tbl3}, our proposed method preserves a good ASR between different models. Such as, the UAP generated by VGG16 can achieve an ASR of over 90\% on VGG19 and over 85\% on ResNet34 and ResNet101. It also can be seen that the generalization ability of the same model architecture is better, such as the perturbation generated by the VGG16 model has a higher ASR in the VGG19 model. It also can be seen that saliency maps has little effect on the generalization ability, which reduce the required PM.

\subsection{Attack selectivity}
The paper \cite{chen2021lie} proposed a soft-threshold defense against adversarial examples method based on attack selectivity. To verify that UAP generated by the proposed method has attack selectivity, we make statistics on the classification results of adversarial examples, as shown in Fig. \ref{fig6}. It can be seen that it will cause attack selectivity when the perturbation is added to the examples. For example, on the VGG19 model, over 88\% of the validation set, which includes 38 classes, is misclassified into three classes. At the same time, under the ResNet34, VGG16, and VGG19 models, the proportion of examples in certain classes has exceeded 80\%, and ResNet101 has exceeded 89\%, equivalent to achieving a targeted attack.

\section{Conclusion}
This paper proposes a UAP generation method for the first time on RSIs. The method utilizes an encoder-decoder network and saliency map to generate the universal perturbation. The former uses the training example to generate the perturbation, while the latter uses the saliency map to modify the perturbation in the sensitive area of the classification model. We apply the universal perturbation attack methods on ordinary images to RSIs and compare it with our proposed method. Experimental results show that our proposed method improves the ASR while reducing the PM. Furthermore, we also verify that it has good generalization ability and attack selectivity.
Our proposed method in generating remote sensing universal perturbation has effectively achieved a high ASR and minor PM. However, the use of a saliency map to modify the perturbation is instability, so we will improve the stability of the saliency map to improve the ASR and PM in the future work.
 
\bibliographystyle{IEEEtran}
\bibliography{refs}


\end{document}